\documentclass{article}

\PassOptionsToPackage{numbers, compress}{natbib}

\usepackage[preprint]{neurips_2025}

\usepackage[utf8]{inputenc} 
\usepackage[T1]{fontenc}    
\usepackage{hyperref}       
\usepackage{url}            
\usepackage{booktabs}       
\usepackage{amsfonts}       
\usepackage{nicefrac}       
\usepackage{microtype}      
\usepackage{xcolor}         

\usepackage{enumitem}
\usepackage{graphicx}
\usepackage{multirow}
\usepackage{makecell}
\usepackage{verbatimbox}
\usepackage{adjustbox}

\usepackage{wrapfig}
\usepackage{tabularx}

\title{A Pilot Empirical Study on \\ When and How to
Use Knowledge Graphs as Retrieval Augmented Generation}

\author{
    \textbf{Xujie Yuan}$^1$ \quad
    \textbf{Yongxu Liu}$^2$ \quad
    \textbf{Shimin Di}$^3$ \quad
    \textbf{Shiwen Wu}$^3$ \quad
    \textbf{Libin Zheng}$^1$ \quad \\
    \textbf{Rui Meng}$^4$ \quad
    \textbf{Lei Chen}$^5$ \quad
    \textbf{Xiaofang Zhou}$^3$ \quad
    \textbf{Jian Yin}$^1$
    \\
    \\
    $^1$SYSU \quad
    $^2$PolyU \quad
    $^3$HKUST \quad
    $^4$BNU-HKBU UIC \quad
    $^5$HKUST(GZ)
}

\begin{document}
\maketitle


\begin{abstract}
The integration of Knowledge Graphs (KGs) into the Retrieval Augmented Generation (RAG) framework has attracted significant interest, with early studies showing promise in mitigating hallucinations and improving model accuracy. However,
a systematic understanding and comparative analysis of the rapidly emerging KG-RAG methods are still lacking.
This paper seeks to lay the foundation for systematically answering the question of when and how to use KG-RAG by analyzing their performance in various application scenarios associated with different technical configurations.
After outlining the mind map using KG-RAG framework and summarizing its popular pipeline, we conduct a pilot empirical study of KG-RAG works to reimplement and evaluate 6 KG-RAG methods across 9 datasets in diverse domains and scenarios, analyzing the impact of 9 KG-RAG configurations in combination with 17 LLMs, and combining Metacognition with KG-RAG as a pilot attempt.
Our results underscore the critical role of appropriate application conditions and optimal configurations of KG-RAG components. 
\end{abstract}

\section{Introduction}
\label{sec:intro}
Recently,
Large Language Models (LLMs) have demonstrated remarkable capabilities in Natural Language Processing (NLP) tasks~\cite{wei2022emergent,10.5555/3495724.3495883}. However, LLMs face critical challenges including hallucination~\cite{sahoo-etal-2024-comprehensive}, limited incorporation with real-time knowledge~\cite{mallen-etal-2023-trust}, and opaque reasoning processes~\cite{Zhou2024larger_and_more}. 
Thus, Retrieval-Augmented Generation (RAG)~\cite{10.5555/3524938.3525306} frameworks have emerged as a promising solution by searching most relevant contents from external knowledge base using similarity methods~\cite{10.1145/3637528.3671470}. 
However, RAG typically treats document contents as independent units, struggling to capture complex relational information and hierarchical interconnections within the data~\cite{liu-etal-2024-lost,li2025structrag}.

To address these limitations, graph-based RAG~\cite{Edge2024FromLT}, particularly those incorporating Knowledge Graphs (KGs) known as KG-RAG, has emerged as a promising paradigm~\cite{zhang-etal-2022-drlk,10.1609/aaai.v38i16.29770,kim-etal-2023-kg,saleh-etal-2024-sg}.
KG-RAG leverages semantic relationships between entities~\cite{li2024subgraphrag} to enable more sophisticated reasoning capabilities~\cite{Sun2023ThinkonGraphDA,wang2025reasoning} and enhance performance in domain-specific applications~\cite{wen-etal-2024-mindmap}.
However, due to the rapid proliferation of related techniques, these KG-RAG works have emerged in a disjointed manner, much like mushrooms after rain, with significant variations in their use of scenarios, datasets, KG-RAG configurations, and LLMs.
They tend to focus on isolated technical innovations across different pipeline stages, without systematic comparison across varied tasks. Moreover, recent reviews~\cite{10387715,Zhang2025ASO,peng2024graphragsurvey,Zhao2024RetrievalAG}
primarily focuses on qualitative analyses, with a lack of quantitative assessments regarding the impact of key configurations across different tasks.

To address this research gap, we aim to explore the key factors that answer the questions of \textit{when} and \textit{how} to use KG-RAG, thereby laying the foundation for a quantitative empirical study. Specifically, we identify two critical gaps in current KG-RAG research: its applicability across diverse scenarios and the effectiveness of different pipeline configurations.
First, the applicability of KG-RAG remains insufficiently explored across several dimensions: task domains (ranging from open-domain to domain-specific tasks), task scenarios (including QA, Diagnosis, and Exams)~\cite{Zhao2024RetrievalAG}, LLM capabilities (from open-source to commercial models), and KG quality (from specialized to general KGs).
Second, the impact of different KG-RAG configurations lacks systematic understanding: (1) pre-retrieval query enhancement strategies (query expansion, decomposition, and understanding), (2) varying retrieval forms (from facts to paths and subgraphs), and (3) post-retrieval prompting approaches (e.g., Chain-of-Thought~\cite{wei2022chain} and Tree-of-Thought~\cite{yao2023tree}).
Through such a systematic investigation, we aim to provide practical guidelines of KG-RAG for answering when and how to use KG-RAG effectively.

In this paper, as a pilot empirical study, we make several contributions to advance the understanding of KG-RAG. We introduce a framework that systematically decomposes KG-RAG into configurable components, enabling controlled analysis. This framework includes our proposed experimental method Pilot, which integrates query understanding with flexible retrieval forms and prompt strategies, achieving substantial performance improvements (e.g., improving accuracy from 61.04\% to 80.16\% on CMB-Exam-Postgraduate). We also pioneer a metacognition-enhanced KG-RAG approach (Meta) as 
an attempt that demonstrates exceptional effectiveness, reaching 98.79\% accuracy on CMB-Exam-Pharmacy, outperforming all commercial models. 
Through experiments, we evaluate 6 KG-RAG methods across 9 datasets and 17 LLMs, employing controls to isolate the effects of 9 different configurations. The results provide quantitative evidence and actionable guidelines for practitioners on when KG-RAG substantially benefits open-source LLMs, precisely which configurations excel in different tasks, and how these enhancements compare to commercial models.

\begin{figure*}[!t]
  \centering
  \includegraphics[width=0.90\linewidth]{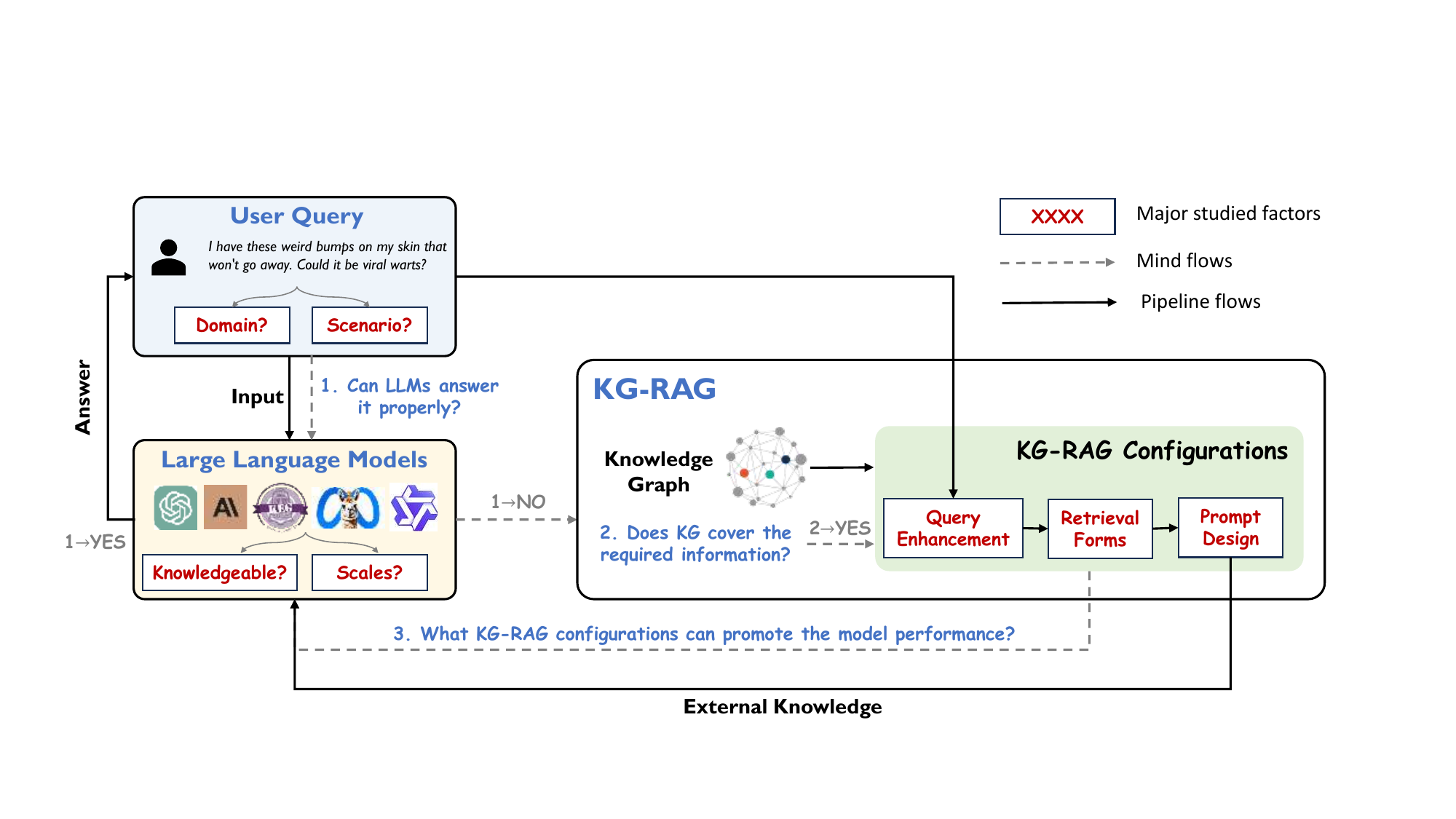}
  \vspace{-5px}
  \caption{The mind and pipeline flows of KG-RAG.}
  \label{fig:mindmap}
  \vspace{-10px}
\end{figure*}

\section{Literature Review}
\label{sec:literature}

\noindent
\textbf{KG-RAG Methods Review.}
Various KG-RAG methods have been proposed in recent years with distinct approaches to utilize KGs. KGRAG~\cite{Soman2023BiomedicalKG} demonstrates a straightforward approach to incorporating KG information by directly retrieving relevant facts and presenting them to LLMs. ToG (Think-on-Graph)~\cite{Sun2023ThinkonGraphDA} introduces a more sophisticated approach that guides LLMs to explore multiple paths within KGs, enables multi-hop reasoning, and provides richer context. MindMap~\cite{wen-etal-2024-mindmap} enhances interpretability by guiding LLMs to construct structured representations that integrate retrieved knowledge from subgraphs. 
RoK~\cite{wang2024reasoning} employs Chain-of-Thought prompts to first extract key entities, then aligns LLMs' pre-trained knowledge with knowledge in KGs. This alignment enables the discovery of more relevant entities. KGGPT~\cite{kim-etal-2023-kg} addresses multi-hop reasoning challenges by breaking down complex queries into simpler clauses, making it easier to construct evidence graphs through separate retrievals for each clause.

\begin{figure}[!t]
    \centering
    \includegraphics[width=0.9\columnwidth]{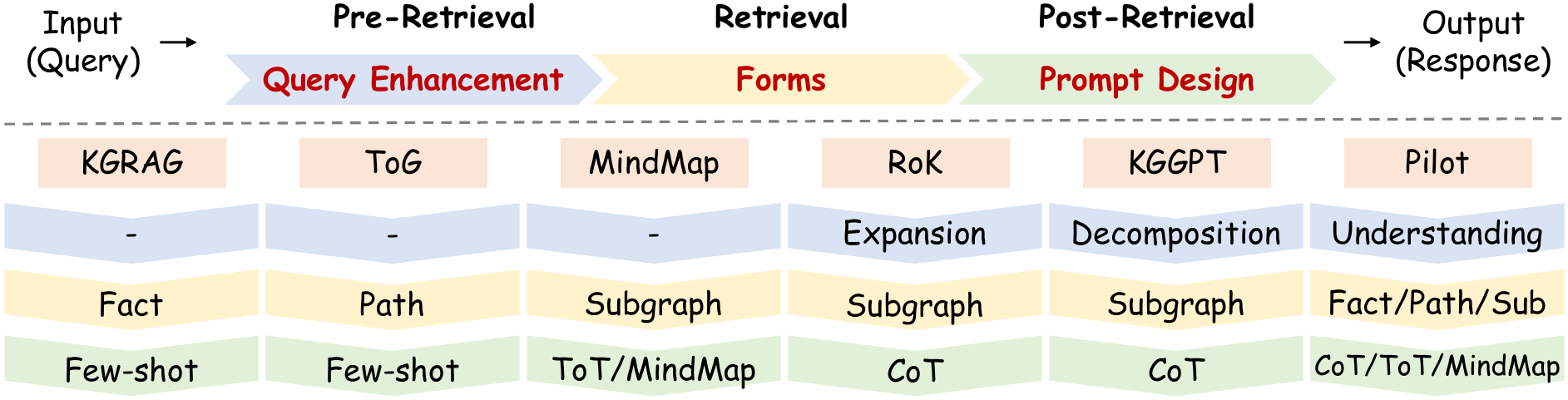}
    \vspace{-5px}
    \caption{KG-RAG Methods and their Configurations.}
    \label{fig_configuration}
    \vspace{-10px}
\end{figure}

While these methods have shown promising results, a systematic comparison of their performance remains limited. As shown in Fig.~\ref{fig_configuration}, these approaches can be categorized by components in three stages: Pre-Retrieval (Query Enhancement), Retrieval (Knowledge Forms), and Post-Retrieval (Prompt Design).
For our empirical study, we also introduce \textbf{Pilot}, an experimental method that combines elements from existing approaches in a configurable framework. It implements query understanding through extraction of key concepts, supports various retrieval forms, and enables different prompting strategies to facilitate systematic ablation studies of individual KG-RAG components.

Recent systematic reviews have provided analyses of RAG frameworks and their integration with KGs~\cite{10387715,Zhao2024RetrievalAG}, establishing a foundation for understanding this rapidly evolving field. CRAG~\cite{yang2024crag} advances the research landscape by introducing a comprehensive benchmark that evaluates RAG performance across multiple dimensions, including domain specificity, data dynamism, content popularity, and question complexity. Complementary research on RAG optimization strategies~\cite{li-etal-2025-enhancing-retrieval} has investigated the impact of various factors on generation quality, such as model size and prompt design. While this study primarily focuses on unstructured text retrieval, its insights provide valuable reference points for KG-RAG.
There are also several surveys that have systematically documented the evolution, technical frameworks, and key components of KG-RAG~\cite{Zhang2025ASO, peng2024graphragsurvey}. These surveys primarily focus on taxonomic classification and theoretical analysis, offering valuable qualitative insights.

\begin{wraptable}{r}{0.7\textwidth}
  \centering
  \vspace{-10px}
  \caption{The statistics of datasets adopted in this paper.}
  \label{tab:data}
  \small
  \setlength\tabcolsep{2pt}
  \begin{tabularx}{\linewidth}{cccccccc}
      \hline
      \textbf{Domain} & \textbf{Dataset} & \textbf{Questions} & \textbf{Language} & \textbf{Scenario} \\
      \hline
        \multirow{3}{*}{Open-domain}
          & CommonsenseQA 
          & 700 
          & English 
          & MC-QA \\ 
          & WebQSP
          & 500 
          & English 
          & KBQA \\ 
          & CWQ
          & 500 
          & English 
          & KBQA \\ 
      \hline
        \multirow{4}{*}{Medical} 
          & GenMedGPT-5K 
          & 700 
          & English 
          & SR-Diagnosis \\
          & CMCQA 
          & 500 
          & Chinese 
          & MR-Diagnosis \\ 
          & CMB-Exam 
          & 3000 
          & Chinese 
          & MC-Exam \\
          & ExplainCPE 
          & 507 
          & Chinese 
          & MC-Exam \\
      \hline
        Legal 
          & JEC-QA 
          & 479
          & Chinese 
          & MC-Exam \\
      \hline
        Multi-domain
          & MMLU
          & 591
          & English 
          & MC-QA \\
      \hline
  \end{tabularx}
  \vspace{-10px}
\end{wraptable}

\noindent
\textbf{Datasets Review.}
KG-RAG methods have been evaluated across a diverse range of datasets, as summarized in Table~\ref{tab:data}.
These datasets span multiple domains and scenarios, reflecting different interaction patterns and information needs that may influence the effectiveness of KG-RAG. CommonsenseQA~\cite{talmor-etal-2019-commonsenseqa} is an open-domain QA dataset focusing on commonsense questions. GenMedGPT-5K~\cite{li2023chatdoctor} and CMCQA~\cite{xia-etal-2022-medconqa} are Domain-specific datasets for clinical consultation, with CMCQA showing higher difficulty through multi-round conversations. CMB-Exam~\cite{wang-etal-2024-cmb} and ExplainCPE~\cite{li-etal-2023-explaincpe} are medical professional examination datasets. 
JEC-QA~\cite{zhong2019jec} is a legal domain dataset derived from Chinese judicial examinations, testing professional legal knowledge through multi-choice questions. MMLU~\cite{hendryckstest2021} extends evaluation to multiple diverse domains, including college biology, high school European history, and professional accounting, enabling assessment of KG-RAG's generalization capabilities. 
More detailed information can be found in Appx.~\ref{appx:exp-data}.

To assess KG-RAG methods across these datasets, we employ multiple evaluation metrics tailored to different task types.
For multiple-choice questions with ground truth answers, we report ``Correct, Wrong, and Fail'' percentages, providing a straightforward measure of accuracy.
For generation tasks with standard reference answers, we employ both lexical and semantic similarity metrics.
ROUGEScore~\cite{lin-2004-rouge} and BERTScore~\cite{bert-score}.
Additionally, we employ G-Eval \cite{liu-etal-2023-g}, an LLM-based evaluation framework that assesses overall answer quality across multiple dimensions including coherence, completeness, and correctness. This provides complementary insights into aspects of response quality beyond simple accuracy or similarity metrics.

\noindent
\textbf{LLMs Review.}
Considering issues such as economic efficiency, open-source availability, and data privacy, researchers have explored whether open-source LLMs (especially low resource consumption) can match or outperform commercial models when enhanced with external knowledge. Consequently, KG-RAG research typically evaluates a range of LLMs with varying scales and architectures.
\begin{itemize}[leftmargin=*]
  \item Qwen1.5-7B~\cite{qwen1.5} and Llama2-7B~\cite{touvron2023llama} are widely used as backbone open-source LLMs (BOS-LLMs) for KG-RAG experiments due to their full open-source status and comparable architectures.
  \item The landscape of open-source models evolves, with other models like Qwen2.5-7B~\cite{yang2024qwen2.5}, Qwen2-72B~\cite{yang2024qwen2}, Deepseek-v2-lite~\cite{Shao2024DeepSeekV2AS}, ChatGLM4-9B \cite{glm2024chatglm}, and Yi-34B~\cite{young2024yi} for Chinese; Llama3.2-1B, Llama3-8B \cite{dubey2024llama}, Llama2-70B, Gemma2-9B \cite{team2024gemma}, and Mixtral-8*7B \cite{jiang2024mixtral} for English, and domain-specialized models like OpenBioLLM-70B~\cite{OpenBioLLMs}.
  \item Commercial LLMs such as Claude3.5-Sonnet\footnote{https://www.anthropic.com/news/claude-3-5-sonnet}, Gemini1.5-Pro\footnote{https://deepmind.google/technologies/gemini/}, GPT4o\footnote{https://openai.com/index/hello-gpt-4o/}, and o1-mini\footnote{https://openai.com/index/openai-o1-mini-advancing-cost-efficient-reasoning/} typically serve as performance benchmarks for comparison.
\end{itemize}

\noindent
\textbf{Research Questions.}
The rapid expansion of KG-RAG research has led to a fragmented landscape. Despite the proliferation of KG-RAG methods across diverse datasets, configurations, and LLMs, there remains limited understanding of their inherent trade-offs across different applications. 
To bridge this gap, we explore the key factors that answer the questions of \textit{when and how to use KG-RAG}, laying the foundation for an empirical study. This paper seeks to answer the following Research Questions (RQs) by conducting a quantitative analysis. 
\begin{itemize}[leftmargin=*]
  \item \textbf{RQ1} (Sec.~\ref{sssec:RQ1}):
  How much do the KG-RAG methods benefit the BOS-LLMs across different tasks?

  \item \textbf{RQ2} (Sec.~\ref{sssec:RQ2}):
  Can KG-RAG enhanced BOS-LLMs offer advantages over commercial LLMs?

  \item \textbf{RQ3} (Sec.~\ref{sssec:RQ3}):
  Which KG-RAG configurations perform best?
\end{itemize}

\section{The Effectiveness of KG-RAG Across Diverse Domains and Scenarios}
\label{sssec:RQ1}
As discussed in Fig.~\ref{fig:mindmap}, answering the question of when to use KG-RAG requires considering whether the LLMs necessitate KG-RAG assistance in specific tasks. In this section, we compare the performance of BOS-LLMs with and without KG-RAG enhancement across the datasets in Tab.~\ref{tab:data}.
We analyze the results from two perspectives~\cite{yang2024crag}, Task Domain (Open vs. Domain-Specific) and Task Scenarios, including Multi-Choice QA (MC-QA), Single-Round Diagnosis (SR-Diagnosis), Multi-Round Diagnosis (MR-Diagnosis), and Multi-Choice Exam (MC-Exam). This analysis provides insights into the conditions under which KG-RAG offers substantial advantages for BOS-LLMs.

\begin{table}[!t]
  \caption{CommonsenseQA (Self-Construted KG)}
  \label{CommonsenseQA}
  \vspace{-5px}
  \centering
  \setlength\tabcolsep{8.2pt}
  \small
  \begin{tabular}{c|ccc|c|ccc}
    \hline
    \multicolumn{4}{c|}{\textbf{LLM only}} & \multicolumn{4}{c}{\textbf{KG-RAG (Llama2-7B)}} \\
    \hline
    \bf Method & \bf Correct & \bf Wrong & \bf Fail & \bf Method & \bf Correct & \bf Wrong & \bf Fail \\ \hline
     Llama2-7B & 39.06 & 60.37 & 0.57 & KGRAG & 42.49 & 56.94 & 0.57 \\
     Llama2-70B & 68.1 & 30.62 & 1.29 & ToG & 42.06 & 57.37 & 0.57 \\
     GPT4o & \bf 84.55 & 15.45 &  0.00 & MindMap & 51.07 & 47.50 & 1.43 \\
     o1-mini & 81.40 & 18.45 & 0.14 & RoK & 42.86 & 57.14 & 0.00 \\
     Claude3.5-S & 82.55 & 17.45 &  0.00 & KGGPT & 48.11 & 51.73 & 0.15 \\
     Gemini1.5-P & 83.83 & 16.17 &  0.00 & Pilot & \underline{51.50} & 48.50 & 0.00 \\
     \hline
  \end{tabular}
  \vspace{-5px}
\end{table}

\begin{table}[!t]
    \caption{GenMedGPT-5K (EMCKG)} 
    \label{GenMedGPT-5K} 
    \vspace{-5px} 
    \centering
    \setlength\tabcolsep{3pt} 
    \small
    \begin{tabular}{c|ccc|cc|c|ccc|cc}
      \hline
      \multicolumn{6}{c|}{\textbf{LLM only}} & \multicolumn{6}{c}{\textbf{KG-RAG (Llama2-7B)}} \\
      \hline
      \textbf{Method} & \textbf{Prec.} & \textbf{Rec.} & \textbf{F1} & \textbf{R-1} & \textbf{R-L} &
      \textbf{Method} & \textbf{Prec.} & \textbf{Rec.} & \textbf{F1} & \textbf{R-1} & \textbf{R-L} \\
      \hline
      Llama2-7B    & 58.79 & 67.89 & 62.96 & 21.02 & 12.21 & KGRAG     & 56.29 & 67.09 & 61.17 & 16.59 & 10.03 \\
      Llama2-70B   & \underline{59.35} & 68.32 & \underline{63.46} & 21.32 & \underline{12.69} & ToG       & 56.50 & 67.80 & 61.59 & 16.93 & 10.06 \\
      GPT4o        & 56.76 & 66.08 & 61.01 & \underline{23.32} & 12.62 & MindMap   & 64.61 & 62.72 & 63.58 & 27.20 & 17.33 \\
      o1-mini      & 58.42 & 57.47 & 57.50 & 17.32 & 10.59 & RoK       & 59.41 & \textbf{71.10} & 64.68 & 23.57 & 14.09 \\
      Claude3.5-S  & 57.01 & \underline{68.35} & 61.29 & 22.37 & 12.01 & KGGPT     & 56.87 & 68.07 & 61.92 & 18.50 & 10.93 \\
      Gemini1.5-P  & 54.49 & 66.50 & 59.87 & 19.07 & 10.24 & Pilot     & \textbf{65.84} & 64.49 & \textbf{65.09} & \textbf{28.49} & \textbf{17.85} \\
      \hline
    \end{tabular}
    \vspace{-5px} 
\end{table}

\begin{table}[!t]
    \caption{CMCQA (CMCKG)}
    \label{CMCQA}
    \vspace{-5px}
    \centering
    \small
    \setlength\tabcolsep{3pt}
    \begin{tabular}{c|ccc|cc|c|ccc|cc}
      \hline
      \multicolumn{6}{c|}{\textbf{LLM only}} & \multicolumn{6}{c}{\textbf{KG-RAG (Qwen1.5-7B)}} \\
      \hline
      \textbf{Method} & \textbf{Prec.} & \textbf{Rec.} & \textbf{F1} & \textbf{R-1} & \textbf{R-L} &
      \textbf{Method} & \textbf{Prec.} & \textbf{Rec.} & \textbf{F1} & \textbf{R-1} & \textbf{R-L} \\
      \hline
      Qwen1.5-7B & 67.61 & 70.57 & 69.00 & 16.75 & 8.91 & KGRAG & 65.65 & 70.01 & 67.71 & \underline{16.45} & \bf 10.58 \\
      Qwen2-72B & 67.50 & 70.35 & 68.84 & 14.94 & 8.17 & ToG & 65.52 & 69.64 & 67.47 & 13.89 & 7.30 \\
      GPT4o & 66.91 & 70.79 & 68.74 & 15.11 & 7.88 & MindMap & 64.93 & 66.14 & 65.46 & 13.51 & 7.83 \\
      o1-mini & 66.24 & 69.07 & 67.55 & 11.03 & 6.08 & RoK & 66.19 & 69.73 & 67.85 & 15.29 & 8.00 \\
      Claude3.5-S & \bf 68.24 & \bf 72.38 & \bf 70.18 & \bf 18.90 & \underline{10.48} & KGGPT & \underline{66.77} & 70.40 & \underline{68.48} & 15.13 & 7.87 \\
      Gemini1.5-P & 67.08 & 70.86 & 68.86 & 12.69 & 6.53 & Pilot & 66.12 & \underline{70.48} & 68.17 & 13.90 & 7.33 \\
      \hline
    \end{tabular}
    \vspace{-10px}
\end{table}

\subsection{Can KG-RAG improve BOS-LLMs across different task domains and scenarios?}
\label{ssssec:RQ1-Analysis}


\textbf{Regarding Task Domains.}
In the Open-Domain task CommonsenseQA (Tab.~\ref{CommonsenseQA}), KG-RAG methods show moderate improvements over the backbone Llama2-7B model. The best performing method (Pilot) achieves a 12.44 percentage improvement (from 39.06\% to 51.50\% accuracy).
In Tab.~\ref{GenMedGPT-5K}, \ref{CMB1}, and \ref{ExplainCPE}, we can also observe that KG-RAG methods deliver significant performance improvements across various Domain-Specific tasks, including diagnoses and Exams. For example, in CMB-Exam-Postgraduate (Tab.~\ref{CMB1}), 
Pilot improves accuracy from 61.04\% to 80.16\%.

\textbf{Regarding Task Scenarios.}
In Single-Round Diagnosis (GenMedGPT-5K, Tab.~\ref{GenMedGPT-5K}), 
the Pilot achieves an F1 score of 65.09,
outperforming the backbone Llama2-7B model (62.96). More significantly, the ROUGE-L improvements are substantial (from 12.21 to 17.85), indicating much better alignment with reference answers. 
The Medical Examinations (CMB-Exam, ExplainCPE) appear particularly well-suited to KG-RAG enhancement. Across all specialties in CMB-Exam (Tab.~\ref{CMB1}), KG-RAG methods provide substantial improvements over the backbone model. 
In ExplainCPE (Tab.~\ref{ExplainCPE}), using KG-RAG methods improves accuracy from 60.08\% to as high as 74.63\%.
In contrast, in multi-round diagnosis scenarios (CMCQA, Tab.~\ref{CMCQA}), KG-RAG methods show limited benefits, with most methods slightly underperforming the base model.

\noindent

\newtheorem{remark}{Remark}[section]
\begin{remark}
\label{remark-3.1}
The effectiveness of KG-RAG exhibits a domain-specificity gradient, moderate improvements in open-domain, but substantial gains in specialized domains. This suggests that KG-RAG's value increases proportionally with domain specialization. Furthermore, we observe a scenario correlation, KG-RAG excels in exams and single-round diagnosis, but struggles with multi-round scenario. This suggests that multi-round dialogues present unique challenges for current KG-RAG approaches, possibly due to challenges in integrating KG information across long context.
\end{remark}

\subsection{More Studies on Generalizability and KG Quality}
\label{sssec:RQ1-Analysis-2-More-Studies}
\noindent
\textbf{Generalizability to Other Domains and Scenarios.}

\begin{wraptable}{r}{0.65\textwidth}
    \centering
    \setlength\tabcolsep{4pt}
    \caption{Results on JEC-QA and MMLU datasets}
    \label{tab:jecqa_mmlu}
    \vspace{2px}
    \begin{tabular}{c|cc|cc}
      \hline
      \multicolumn{1}{c|}{\bf Datasets} 
        & \multicolumn{2}{c|}{\bf JEC-QA} 
        & \multicolumn{2}{c}{\bf MMLU (HSEH)} \\ 
      \hline
        & \bf Method & \bf Correct 
        & \bf Method & \bf Correct \\ 
      \hline
      BOS-LLM 
        & Qwen2.5-7B & 65.06 
        & Llama2-70B & 71.95 \\ 
      \hline
      \multirow{2}{*}{KG-RAG} 
        & MindMap    & 71.55 
        & MindMap    & 70.12 \\ 
        & Pilot      & 72.80 
        & Pilot      & 75.00 \\ 
      \hline
    \end{tabular}
    \vspace{-5px}
\end{wraptable}

To examine the generalizability, we expanded additional datasets, JEC-QA and MMLU. Our experiments span diverse domains, 4 medical datasets discussed earlier, 1 legal domain dataset (JEC-QA), and 3 specialized subjects from MMLU (High School European History, College Biology, and Professional Accounting). 
The results in Tab.~\ref{tab:jecqa_mmlu} demonstrate that KG-RAG methods remain effective in specialized domains beyond medical. In the legal domain, both MindMap and Pilot improved accuracy over the base LLMs. On MMLU (High School European History, HSEH), the Pilot improved accuracy from 71.95\% to 75.00\%. For results on other MMLU domains, please refer to Appx.~\ref{appx:MMLU}.
To further understand KG-RAG capabilities in structured knowledge-intensive tasks, we explored performance on dedicated Knowledge Base Question Answering (KBQA) datasets WebQSP~\cite{Yih2016TheVO} and Complex Web Questions (CWQ)~\cite{talmor-berant-2018-web} in Appx.\ref{KBQA-Results}.

\noindent
\textbf{Impact of Knowledge Graph Quality.}
Following~\cite{wen-etal-2024-mindmap}, we utilize EMCKG and CMCKG as the KGs for GenMedGPT-5 and CMCQA, respectively. Besides, we construct the corresponding KGs for the remaining datasets
(detailed in Appx.~\ref{appx:KG-Construction}). 
To examine the impact of KG quality~\cite{Sui2024CanKG}, we conducted experiments on the ExplainCPE using spKG (specialized KG) and CMCKG (only partially covers the required knowledge) in Tab.~\ref{ExplainCPE}. Using a specialized KG led to substantial performance improvements (from 60.08\% to 73.26\%), highlighting how KG completeness critically affect KG-RAG effectiveness.

\begin{table}[!t]
  \centering
  \setlength\tabcolsep{2pt}
  \caption{CMB under Postgraduate, Nursing and Pharmacy (Self-Constructed KG)}
  \label{CMB1}
  \vspace{-5px}
  \begin{tabular}{c|c|ccc|ccc|ccc}
    \hline
    \multicolumn{10}{c}{\bf CMB} \\ 
    \hline
    \multicolumn{2}{c|}{} 
      & \multicolumn{3}{c|}{\bf Postgraduate} 
      & \multicolumn{3}{c|}{\bf Nursing} 
      & \multicolumn{3}{c}{\bf Pharmacy} \\ 
    \hline
    \bf Type & \bf Method 
      & \bf Correct & \bf Wrong & \bf Fail 
      & \bf Correct & \bf Wrong & \bf Fail 
      & \bf Correct & \bf Wrong & \bf Fail \\ 
    \hline
    \multirow{6}{*}{\makecell{LLM\\only}} 
      & Qwen1.5-7B    & 61.04 & 36.14 & 2.81  & 75.55 & 23.65 & 0.80  & 63.93 & 36.07 & 0.00 \\ 
      & Qwen2-72B     & \underline{87.78} & 12.02 & 0.20 & \underline{89.78} & 10.22 & 0.00 & \underline{90.18} & 9.82 & 0.00 \\ 
      & GPT4o         & 76.95 & 22.44 & 0.60  & 83.13 & 16.87 & 0.00  & 72.89 & 26.91 & 0.20 \\ 
      & o1-mini       & 63.13 & 36.27 & 0.60  & 74.50 & 25.50 & 0.00  & 60.44 & 39.56 & 0.00 \\ 
      & Claude3.5-S   & 69.54 & 30.46 & 0.00  & 75.90 & 24.10 & 0.00  & 65.86 & 34.14 & 0.00 \\ 
      & Gemini1.5-P   & 75.95 & 24.05 & 0.00  & 80.72 & 19.28 & 0.00  & 70.68 & 29.32 & 0.00 \\ 
    \hline
    \multirow{6}{*}{\makecell{KG-RAG\\(Qwen1.5-7B)}} 
      & KGRAG         & 76.05 & 21.64 & 2.31  & 84.82 & 14.35 & 0.83  & 76.53 & 21.84 & 1.63 \\ 
      & ToG           & 72.75 & 27.05 & 0.20  & 75.75 & 23.85 & 0.40  & 71.14 & 28.26 & 0.60 \\ 
      & MindMap       & 78.11 & 21.49 & 0.40  & 82.57 & 17.23 & 0.20  & 76.75 & 22.65 & 0.60 \\ 
      & RoK           & 72.73 & 27.27 & 0.00  & 85.21 & 14.79 & 0.00  & \underline{77.78} & 22.22 & 0.00 \\ 
      & KGGPT         & 70.99 & 29.01 & 0.00  & 75.60 & 24.40 & 0.00  & 68.40 & 31.60 & 0.00 \\ 
      & Pilot         & \underline{80.16} & 19.44 & 0.40  & \underline{85.37} & 14.43 & 0.20  & 77.35 & 22.04 & 0.60 \\ 
    \hline
    \multirow{2}{*}{\makecell{KG-RAG\\(Qwen2-72B)}} 
      & Pilot         & 88.98 & 11.02 & 0.00  & 91.98 &  8.02 & 0.00  & 92.18 &  7.82 & 0.00 \\ 
      & Meta          & \bf92.99 & 7.01 & 0.00  & \bf96.54 & 3.46 & 0.00  & \bf98.79 & 1.21 & 0.00 \\ 
    \hline
  \end{tabular}
  \vspace{-10px}
\end{table}

\begin{table}[!b]
    \centering
    \setlength\tabcolsep{4.5pt}
    \caption{ExplainCPE (BOS-LLMs: Qwen1.5-7B, Qwen2.5-7B, and Qwen2-72B)}
    \label{ExplainCPE}
    \vspace{-5px}
    \begin{tabular}{c|c|c|c|c|c|c}
      \hline
      \multicolumn{2}{c|}{\textbf{LLM}} & \multirow{2}{*}{\textbf{KG-RAG}} & \textbf{spKG} & \textbf{CMCKG} & \textbf{spKG} & \textbf{spKG} \\
      \multicolumn{2}{c|}{\textbf{Only}} & & \textbf{Qwen1.5-7B} & \textbf{Qwen1.5-7B} & \textbf{Qwen2.5-7B} & \textbf{Qwen2-72B} \\
      \hline
      \textbf{Method} & \textbf{Correct} & \textbf{Method} & \textbf{Correct} & \textbf{Correct} & \textbf{Correct} & \textbf{Correct} \\
      \hline
      Qwen1.5-7B   & 60.08 & KGRAG    & 69.88 & 58.22 & 78.53 & 87.93 \\
      Qwen2-72B    & \underline{81.82} & ToG      & 68.58 & \underline{61.07} & 78.85 & 83.60 \\
      GPT4o         & 79.64 & MindMap  & 70.68 & 56.92 & 78.41 & 87.98 \\
      o1-mini      & 75.10 & RoK      & \underline{74.63} & 58.29 & \underline{79.12} & \textbf{88.06} \\
      Claude3.5-S  & 76.88 & KGGPT    & 63.69 & 53.00 & 78.86 & 87.15 \\
      Gemini1.5-P  & 69.37 & Pilot    & 73.26 & 55.93 & 77.89 & 87.93 \\
      \hline
    \end{tabular}
    \vspace{-10px}
  \end{table}

\section{Bridging the Gap: KG-RAG Enhanced BOS-LLMs vs. Commercial Models}
\label{sssec:RQ2}
Beyond the task domains and scenarios, the capability of LLMs is also a key factor in determining the effectiveness of KG-RAG. This analysis is important for practical applications, where considerations of cost, privacy, and deployment constraints may make BOS-LLMs preferable compared with commercial models if their performance can be sufficiently enhanced. 

\subsection{Can BOS-LLMs with KG-RAG are better than commercial LLMs?}
\textbf{Regarding task domains.}
In Open-Domain task CommonsenseQA (Tab.~\ref{CommonsenseQA}), KG-RAG methods show moderate improvements over the backbone Llama2-7B model. However, this improvement is limited compared to commercial models (GPT4o, 84.55\%), as commercial LLMs may have already internalized sufficient commonsense knowledge during pre-training.
In Domain-Specific tasks, BOS-LLMs with KG-RAG methods can match or even surpass commercial LLMs (Tab.~\ref{GenMedGPT-5K}, \ref{CMB1}, and \ref{ExplainCPE}). For example, in CMB-Exam-Nursing (Tab.~\ref{CMB1}), Pilot (85.37\%) outperforms GPT-4o (83.14\%). These results indicate that, in domain-specific tasks, given the economic advantages of BOS-LLMs over commercial LLMs, KG-RAG plays a significant role and remains valuable.

\textbf{Regarding task scenarios.}
In Single-Round Diagnosis (GenMedGPT-5K, Tab.~\ref{GenMedGPT-5K}), the Pilot achieves F1 of 65.09, outperforming not only the backbone model Llama2-7B but also commercial models like Claude3.5-S (61.29). The ROUGE-L shows even more dramatic improvement, with Pilot (17.85) outperforming all commercial models.
In Medical Examinations (CMB-Exam and ExplainCPE), KG-RAG methods show strong competitiveness. For instance, in CMB-Exam-Pharmacy (Tab.~\ref{CMB1}), Pilot achieves 77.35\%  accuracy, surpassing GPT4o (72.89\%). These results indicate that for examination scenarios, KG-RAG narrows the gap with commercial models.
However, in Multi-Round Diagnosis (CMCQA, Tab.~\ref{CMCQA}), we observe that commercial models maintain advantages. Claude3.5-S achieves the highest F1 (70.18). The best KG-RAG method KGGPT (F1 68.48) slightly underperformed.
This suggests that for conversational scenarios, commercial LLMs retain advantages in generating coherent responses, an area where current KG-RAG approaches show limitations.

\subsection{KG-RAG with Advanced BOS-LLMs}
To investigate whether KG-RAG benefits persist with newer and larger LLMs, we expanded experiments with Qwen2.5-7B and Qwen2-72B as BOS-LLMs. 
As shown in Tab~\ref{ExplainCPE} (ExplainCPE), with Qwen2.5-7B, KG-RAG (RoK) continue improves accuracy to 79.12\%, exceeding Claude3.5-S (76.88\%). With Qwen2-72B, RoK improves accuracy from 81.82\% to 88.06\%, outperforming all commercial models. In CMB-Exam (Tab.~\ref{CMB1}), Pilot further enhances Qwen2-72B's strong performance across all specialties (e.g., from 89.78\% to 91.98\% in Nursing). 
While the magnitude of improvement generally decreases as model size increases, the consistent gains across all tested models demonstrate that external knowledge integration remains valuable even as LLM capabilities evolve. This suggests that KG-RAG provides complementary benefits that persist across model generations and scales.

\noindent
\textbf{Remark 4 } \textit{KG-RAG provides a comparative advantage in specialized domains, where even commercial LLMs may lack sufficient domain knowledge. KG-RAG's effectiveness relative to commercial models varies by scenarios, with better performance in examination and diagnostic tasks but limitations in conversational tasks. KG-RAG benefits remain consistent across model generations and scales. This suggests that KG-RAG provides complementary knowledge that remains valuable even as internal model knowledge improves.}

\section{Optimizing KG-RAG Performance: Component Configurations}
\label{sssec:RQ3}
Previous sections have established \textit{\textbf{when}} KG-RAG provides meaningful advantages, primarily in domain-specific tasks including single-round diagnosis and examinations, while offering limited benefits in open domains and conversational scenarios.
Considering KG-RAG adoption, \textit{\textbf{how}} to optimally configure KG-RAG becomes a question.  
Since implementing KG-RAG requires engineering effort, understanding which configurations work best is essential for maximizing performance benefits.

\subsection{KG-RAG Configurations}
As discussed in Sec.~\ref{sec:literature} (shown in Fig.~\ref{fig_configuration}), we review five existing KG-RAG works and summarize three main modules based on the stage, Pre-Retrieval, Retrieval, and Post-Retrieval.
Additionally, to facilitate subsequent ablation experiments for validating modules, we supplement an experimental Pilot method, as proposed in this paper.

\textbf{Query Enhancement in Pre-retrieval.}
The Pre-Retrieval phase focuses on determining ``what to retrieve'' by aligning queries with knowledge base content~\cite{jiang2024hykgehypothesisknowledgegraph}. RoK's \textit{\textbf{Query Expansion}}~\cite{wang2024reasoning}, which leverages Chain-of-Thought to extract key entities. KGGPT's \textit{\textbf{Query Decomposition}}~\cite{kim-etal-2023-kg}, which breaks complex queries into simpler clauses for targeted retrieval. \textit{\textbf{Query Understanding}} in our Pilot method, which extracts the main ideas from queries to ensure alignment with its intention~\cite{Gan2024SimilarityIN}.


\textbf{Retrieval Forms After Retrieval.}
In the retrieval phase, KG-RAG organizes retrieved graph context that can be input to LLMs as reference. Due to differences in retrieval mechanisms, the graph context may be organized into three forms (fact, path, and subgraph) with increasing information granularity.
\textit{\textbf{Fact}} is the most basic knowledge unit in triplet form \texttt{(Sub,Pred,Obj)}, providing discrete, structured knowledge points~\cite{Soman2023BiomedicalKG}, while precise, lack contextual connections. \textit{\textbf{Path}} consists of connected triplet sequences. ToG~\cite{Sun2023ThinkonGraphDA} supports multi-hop reasoning by guiding LLMs to explore multiple paths. Paths can balance information density with structural clarity, but may miss broader relationships outside the path. \textit{\textbf{Subgraph}} combines both paths and neighboring entities, and can capture more comprehensive patterns, enabling a more thorough understanding in greater detail. MindMap~\cite{wen-etal-2024-mindmap} employs both path-based and neighbor-based exploration, ultimately combining path and neighbor information to form an evidence subgraph.


\textbf{Prompt Design After Retrieval.}
In the Post-Retrieval phase, while some works focus on filtering~\cite{li-etal-2024-llatrieval} or reranking~\cite{glass-etal-2022-re2g} retrieved results, we primarily investigate how different prompts guide the reasoning process with retrieved knowledge~\cite{Sahoo2024ASS,Tonmoy2024ACS,Chen2023UnleashingTP}. We mainly examine three following prompt patterns.
\textit{\textbf{Chain-of-Thought (CoT)}} introduces step-by-step reasoning~\cite{wei2022chain}, breaking complex problems into sequential intermediate steps. 
\textit{\textbf{Tree-of-Thought (ToT)}}~\cite{yao2023tree} extends this concept by enabling multi-branch exploration, allowing LLMs to simultaneously consider and compare multiple reasoning paths. 
\textit{\textbf{MindMap}}~\cite{wen-etal-2024-mindmap} enhances reasoning interpretability by guiding LLMs to construct structured mind maps that integrate retrieved knowledge while maintaining reasoning traces.


\subsection{How effective are different KG-RAG configurations?}

\begin{table}[!t]
  \centering
  \caption{Pre-Retrieval Query enhancement results}
  \label{pre-Ablation}
  \vspace{-5px}
  \small
  \setlength\tabcolsep{4.5pt}
  \begin{tabular}{l|l|c|ccc|cc}
  \hline
  \multirow{2}{*}{Datasets} & \multirow{2}{*}{Methods} & \multirow{2}{*}{\centering Acc} & \multicolumn{3}{c|}{BERT Score} & \multicolumn{2}{c}{ROUGE Score} \\
  & & & Precision & Recall & F1 & ROUGE-1 & ROUGE-L \\
  \hline
  \multirow{4}{*}{GenMedGPT-5K} 
  & w/o Enhancement & - & 0.6499 & 0.6402 & 0.6443 & \textbf{0.2901} & \textbf{0.1802} \\
  & Understand (Pilot) & - & \textbf{0.6584} & 0.6449 & \textbf{0.6509} & 0.2849 & 0.1785 \\
  & Expanse (RoK) & - & 0.5941 & \textbf{0.7110} & 0.6468 & 0.2357 & 0.1409 \\
  & Decompose (KGGPT) & - & 0.5687 & 0.6807 & 0.6192 & 0.1850 & 0.1093 \\
  \hline
  \multirow{4}{*}{ExplainCPE}
  & w/o Enhancement & 66.80 & 0.7279 & 0.7537 & 0.7354 & 0.3020 & 0.1963 \\
  & Understand (Pilot) & 73.26 & \textbf{0.7281} & 0.7515 & 0.7337 & 0.3000 & 0.1950 \\
  & Expanse (RoK) & \textbf{74.63} & 0.7242 & \textbf{0.7670} & \textbf{0.7439} & 0.2961 & 0.1973 \\
  & Decompose (KGGPT) & 63.69 & 0.7223 & 0.7638 & 0.7414 & \textbf{0.3169} & \textbf{0.2103} \\
  \hline
  \end{tabular}
  \vspace{-10px}
\end{table}

\begin{table}[!b]
  \centering
  \small
  \caption{Configurations comparison on GenMedGPT-5K}
  \label{GenMedGPT-Config}
  \vspace{-5px}
  \setlength\tabcolsep{5pt}
  \begin{tabular}{l|ccc|ccc|cccc}
  \hline
  \multirow{2}{*}{Config} & \multicolumn{3}{c|}{BERT Score} & \multicolumn{3}{c|}{ROUGE Score} & \multicolumn{4}{c}{G-Eval} \\
  & Prec. & Rec. & F1 & R-1 & R-2 & R-L & CR & Comp & Corr & Emp \\
  \hline
  Facts\_w/o Prompt & 56.62 & \underline{67.74} & 61.64 & 17.16 & 3.44 & 10.31 & \underline{99.95} & \underline{97.83} & \underline{99.24} & \underline{79.29} \\
  Facts+CoT & 58.51 & 59.53 & 59.00 & 25.32 & 4.91 & 14.70 & 99.77 & 87.44 & 97.36 & 78.22 \\
  Facts+ToT & 64.42 & 63.43 & 63.91 & 28.54 & 5.86 & 17.53 & 79.30 & 62.11 & 69.86 & 55.07 \\
  Facts+MindMap & \underline{65.10} & 63.78 & \underline{64.37} & \textbf{28.70} & \underline{6.02} & \underline{17.82} & 99.49 & 82.10 & 96.62 & 77.64 \\
  \hline
  Path\_w/o Prompt & 56.85 & \textbf{68.00} & 61.89 & 18.32 & 3.77 & 10.89 & {\textbf{100.00}} & {\textbf{97.93}} & {\textbf{99.29}} & {\textbf{79.44}} \\
  Path+CoT & 58.02 & 59.01 & 58.50 & 25.05 & 4.82 & 14.53 & 99.91 & 87.09 & 98.83 & 79.28 \\
  Path+ToT & 63.85 & 62.92 & 63.41 & 28.42 & 5.75 & 17.43 & 83.00 & 65.60 & 77.00 & 60.00 \\
  Path+MindMap & \textbf{65.84} & 64.49 & \textbf{65.09} & \underline{28.49} & \textbf{6.07} & \textbf{17.85} & 99.54 & 81.33 & 97.56 & 78.10 \\
  \hline
  Subgraph\_w/o Prompt & 56.40 & \underline{67.01} & \underline{61.21} & 16.84 & 3.26 & 10.30 & 98.43 & \underline{94.39} & \underline{97.93} & \underline{78.08} \\
  Subgraph+CoT & 58.49 & 61.43 & 59.85 & 25.21 & 4.72 & 14.47 & \underline{99.47} & 88.67 & 96.76 & 77.77 \\
  Subgraph+ToT & 57.83 & 59.92 & 58.94 & 25.38 & 4.96 & 14.92 & 75.32 & 57.91 & 65.27 & 51.24 \\
  Subgraph+MindMap & \underline{59.29} & 58.01 & 58.60 & \underline{26.16} & \underline{5.57} & \underline{16.13} & 97.13 & 79.21 & 92.42 & 74.44 \\
  \hline
  \end{tabular}
  \vspace{-10px}
\end{table}

In this subsection, we compare the performance of different KG-RAG configurations in Tab.~\ref{pre-Ablation}, \ref{GenMedGPT-Config}, and \ref{ExplainCPE-Config}. For our analysis, we focus on GenMedGPT-5K and ExplainCPE, where KG-RAG showed substantial improvements. Although previous results indicate that KG-RAG may not provide benefits for CMCQA, we deliberately include its configuration analysis to identify potential limitations and improvement directions in Appx.~\ref{appendix:CMCQA-config}.

\textbf{Impact of Query Enhancement.}
In Tab.~\ref{pre-Ablation}, we compare the impact of query enhancement methods.
Our analysis reveals that optimal query enhancement strategies vary significantly based on task characteristics.
Understanding (Pilot) works best for medical diagnosis (GenMedGPT-5K, F1 0.6509). Expansion (RoK) performs strongest for exams (ExplainCPE, 74.63\% accuracy). 
These differences suggest that the nature of the task should guide configuration selection. Diagnosis benefits from precise symptom identification, while examinations require broader conceptual connections to address knowledge requirements. We further provide a case study in the Appx.~\ref{appx:case-study} to support our analysis.

\textbf{Impact of Retrieval Forms.}
In Tab.~\ref{GenMedGPT-Config} and \ref{ExplainCPE-Config}, we compare the impact of retrieval forms, including fact, path, and subgraph.
On GenMedGPT-5K (Tab.~\ref{GenMedGPT-Config}), using facts and paths typically outperforms subgraphs in terms of BERT and ROUGE Scores. Similarly, using facts shows better performance on ExplainCPE (Tab.~\ref{ExplainCPE-Config}). Interestingly, when examining by G-Eval, the differences between retrieval forms become minimal when prompts are not used. This suggests that while retrieval form affects linguistic metrics, the factual quality of answers may depend more on whether retrieved knowledge contains the necessary information. 

\textbf{Impact of Prompt Design.}
In Tab.~\ref{GenMedGPT-Config} and \ref{ExplainCPE-Config}, we also compare the impact of prompt strategies, including CoT, ToT, and MindMap.
Counterintuitively, across both tasks, configurations without prompts outperform prompted strategies in G-Eval metrics. However, prompting strategies (especially MindMap) perform better in BERT and ROUGE Scores. 
This divergence suggests that direct knowledge utilization without elaborate prompting produces more factually relevant answers, while prompting strategies may improve language quality at the cost of conciseness and factual density.

\textbf{Configuration Interactions.}
To understand how different KG-RAG components influence each other, we also analyzed configuration interactions.
As shown in Tab.~\ref{GenMedGPT-Config}, Path excels in ROUGE-L by providing causal reasoning chains. Path+MindMap outperforms Path+ToT because MindMap's structured approach better handles the continuity needed for reasoning than ToT's branching exploration.
Regarding synergistic effects between query enhancement and retrieval forms. When using the Pilot (Query Understanding) with ExplainCPE (Tab.~\ref{ExplainCPE-Config}), performance improved dramatically with Fact but showed less improvement with Path. 
These patterns suggest that optimal configuration isn't about selecting the best individual components, but rather identifying synergistic combinations.


\textbf{A Pilot Attempt at Metacognition-enhanced KG-RAG.}
In CMB-Exam, we introduce a Metacognition-enhanced~\cite{wang-zhao-2024-metacognitive, zhou2024metacognitive} approach (Meta) that incorporates backtracking and self-correction mechanisms that address error accumulation and local optimality issues in the linear retrieval process. As shown in Tab.~\ref{CMB1}, this approach achieves exceptional results in Pharmacy (98.79\% accuracy).

\textbf{Remark 5.2 } \textit{
\begin{itemize}[leftmargin=*]
\item For Query Enhancement, understanding excels in diagnoses, while expansion performs better for exams. This pattern emerges because diagnostic tasks require high precision, and understanding methods maintain accuracy by focusing on core symptoms. Conversely, examination benefits from expansion methods that connect symptoms with related conditions and treatments. 
\item For Retrieval Forms, the surprising discovery is that simpler forms (facts and paths) consistently outperform subgraphs. This challenges the intuition that more comprehensive knowledge representation necessarily leads to better performance, indicating that focused, relevant knowledge may be more valuable than extensive but potentially noisy subgraphs.  
\item Regarding Prompt Design, without elaborate prompting often produce higher-quality answers according to G-Eval, despite scoring lower on linguistic measures. This suggests that direct knowledge utilization may preserve factual accuracy better than complex reasoning frameworks. 
\end{itemize}
Overall, task characteristics determine optimal configurations. Components' performance is not isolated but emerges from their combinations. This implies that KG-RAG optimization should adopt a holistic approach tailored to specific task requirements.
}

\begin{table}[!t]
  \centering
  \small
  \caption{Configurations comparison on ExplainCPE}
  \label{ExplainCPE-Config}
  \vspace{-5px}
  \setlength\tabcolsep{3.8pt}
  \begin{tabular}{l|c|ccc|ccc|cccc}
  \hline
  \multirow{2}{*}{Config}   & \multirow{2}{*}{\centering Acc} & \multicolumn{3}{c|}{BERT Score} & \multicolumn{3}{c|}{ROUGE Score} & \multicolumn{4}{c}{G-Eval} \\
  & & Prec. & Rec. & F1 & R-1 & R-2 & R-L & CR & Comp & Corr & Emp \\
  \hline
  Facts\_w/o Prompt & {\textbf{73.26}} & \underline{72.81} & 75.15 & 73.37 & {\textbf{30.00}} & {\textbf{9.47}} & {\textbf{19.50}} & {\textbf{95.83}} & {\textbf{90.87}} & {\textbf{92.27}} & {\textbf{86.28}} 
  \\
  Facts+CoT & 69.83 & 71.20 & {\textbf{78.33}} & {\textbf{74.52}} & 22.35 & 7.30 & 14.74 & 79.80 & 79.94 & 79.74 & 80.30 \\
  Facts+ToT & 65.91 & 68.92 & 77.71 & 72.98 & 16.56 & 5.31 & 10.89 & 79.50 & 79.86 & 80.06 & 79.94 \\
  Facts+MindMap & 59.50 & 67.12 & 75.22 & 70.85 & 15.43 & 5.09 & 10.30 & 79.59 & 79.51 & 80.01 & 79.53 \\
  \hline
  Path\_w/o Prompt & \underline{63.22} & 72.96 & 69.83 & 69.43 & \underline{26.50} & \underline{8.11} & \underline{17.21} & \underline{94.02} & \underline{84.45} & \underline{91.09} & \underline{82.68} \\
  Path+CoT & 58.68 & {\textbf{76.11}} & \underline{77.46} & \underline{74.12} & 21.14 & 7.00 & 14.40 & 79.75 & 79.89 & 79.95 & 79.94 \\
  Path+ToT & 56.20 & {\textbf{76.11}} & 77.10 & 73.89 & 20.57 & 6.99 & 14.22 & 79.82 & 79.59 & 79.75 & 79.83 \\
  Path+MindMap & 55.37 & 67.07 & 75.24 & 70.84 & 14.97 & 4.80 & 9.97 & 80.08 & 80.04 & 80.34 & 79.64 \\
  \hline
  Subgraph\_w/o Prompt & \underline{66.74} & \underline{71.06} & 63.91 & 65.14 & 15.62 & 6.01 & 11.96 & \underline{94.97} & \underline{84.00} & \underline{91.79} & \underline{82.56} \\
  Subgraph+CoT & 63.22 & \underline{71.06} & \underline{78.32} & \underline{74.44} & \underline{21.87} & \underline{7.11} & \underline{14.47} & 80.30 & 80.27 & 80.06 & 80.53 \\
  Subgraph+ToT & 61.16 & 68.89 & 77.70 & 72.97 & 16.46 & 5.27 & 10.88 & 80.06 & 79.72 & 79.93 & 80.16 \\
  Subgraph+MindMap & 56.20 & 67.14 & 75.18 & 70.84 & 15.36 & 4.96 & 10.21 & 80.50 & 80.38 & 79.79 & 80.23 \\
  \hline
  \end{tabular}
  \vspace{-10px}
\end{table}

\section{Conclusion}
This study systematically explores the applicability and configurations of KG-RAG across different tasks. 
Our findings demonstrate that KG-RAG enhanced BOS-LLMs can match or surpass commercial models in domain-specific tasks (particularly single-round diagnosis and examinations) with benefits persisting across model scales, while offering limited benefits in open domain and conversational scenario. Configurations should be task-driven. Query Understanding excels in diagnosis, while Expansion performs best in exams. Fact and Path outperform Subgraph for concise questions, and surprisingly, prompting in the post-retrieval stage is actually not very important. Our findings provide a practical framework for determining when and how to implement KG-RAG effectively, enabling practitioners to maximize benefits while avoiding unnecessary costs. 
Despite comprehensive empirical study, this work still has several limitations discussed in Appx.~\ref{appendix:limitations} that warrant further investigation.




\bibliographystyle{plainnat}
\bibliography{biblio}

\appendix

\section{Complementary Experiments}
\label{appx:exp}

\subsection{Datasets}
\label{appx:exp-data}

The detailed descriptions of the adopted KG-RAG datasets are summarized as follows:
\begin{itemize}[leftmargin=*]
  \item CommonsenseQA~\cite{talmor-etal-2019-commonsenseqa} is a multiple-choice QA dataset specifically designed to evaluate commonsense reasoning capabilities. 
  Each question is accompanied by five candidate answers, only one of which is correct.
  
  \item GenMedGPT-5K~\cite{li2023chatdoctor} is a medical dialogue dataset, covers patient-doctor single round dialogues. Generated through interactions between GPT-3.5 and the iCliniq disease database, this dataset contains clinical conversations covering patient symptoms, diagnoses, recommended treatments, and diagnostic tests.
  
  \item CMCQA~\cite{xia-etal-2022-medconqa} is a comprehensive medical conversational QA dataset derived from professional Chinese medical consultation platform. The dataset encompasses multi-round clinical dialogues across 45 medical specialties, including andrology, stomatology, and obstetrics-gynecology, representing diverse clinical interactions between healthcare providers and patients. 
  
  \item
  CMB-Exam~\cite{wang-etal-2024-cmb} covers 280,839 questions from six major medical professional qualification examinations, including physicians, nurses, medical technologists and pharmacists, as well as Undergraduate Disciplines Examinations and Graduate Entrance Examination in the medical field at China.
  Given the extensive scale of CMB-Exam, we sample a subset of CMB-Exam that comprise 3,000 questions, where 500 questions are randomly sampled from each category.
  
  \item ExplainCPE~\cite{li-etal-2023-explaincpe} is a Chinese medical benchmark dataset containing over 7K instances from the National Licensed Pharmacist Examination. This dataset is distinctive in providing both multiple-choice answers and their corresponding explanations.

  \item JEC-QA~\cite{zhong2019jec} is a legal domain dataset collected from the National Judicial Examination of China. It serves as a comprehensive evaluation of professional skills required for legal practitioners. The dataset is particularly challenging as it requires logical reasoning abilities to retrieve relevant materials and answer questions correctly.

  \item MMLU~\cite{hendryckstest2021} is a multitask benchmark across 57 diverse tasks spanning elementary mathematics, US history, computer science, law, and other domains. The benchmark requires extensive world knowledge and problem-solving abilities to achieve high accuracy. MMLU comprehensively assesses both breadth and depth of a model's academic and professional understanding.
\end{itemize}

Additionally, we incorporated two representative KBQA datasets, WebQSP \cite{Yih2016TheVO} and Complex Web Questions (CWQ) \cite{talmor-berant-2018-web}, discussing KBQA as a special case. WebQSP consists of natural language questions emphasizing single-hop factoid queries, while CWQ features more complex multi-hop questions requiring compositional reasoning over knowledge graphs.

\subsection{The remaining Experimental Results}

\begin{table}[!t]
  \centering
  \small
  \vspace{-5px}
  \setlength\tabcolsep{3pt}
  \caption{Experimental Results on KBQA Datasets}
  \label{KBQA-Table}
  \begin{tabular}{llcc}
  \hline
  \textbf{Type} & \textbf{Method} & \textbf{WebQSP} & \textbf{CWQ} \\
  \hline
  \multirow{2}{*}{KG-RAG} & MindMap (Llama2-7B) & 30.82 & 30.51 \\
  & ToG (ChatGPT)~\cite{Sun2023ThinkonGraphDA} & \textbf{75.80} & \textbf{58.90} \\
  \hline
  \multirow{5}{*}{LLM only} & Llama3.2-1B & 37.45 & 15.21 \\
  & Llama2-7B & 43.29 & 21.91 \\
  & Llama3-8B & 55.41 & 28.09 \\
  & Llama2-70B & 53.68 & 28.87 \\
  & Mixtral-8*7B & 58.01 & 33.25 \\
  & ChatGPT & \underline{63.30} & \underline{37.60} \\
  \hline
  \end{tabular}
  \vspace{-10px}
\end{table}

\subsubsection{KBQA experimental results}
\label{KBQA-Results}
We conducted experiments on two KBQA datasets and the results are shown in Tab.~\ref{KBQA-Table}. The results demonstrate significant performance variations across different LLMs and KG-RAG methods, highlighting the importance of systematic understanding when deploying KG-RAG in knowledge-intensive tasks.

\subsubsection{Remaining results of CMB-Exam}
\label{appx:CMB-Exam} 
Due to space constraints, the remaining experimental results of the CMB-Exam (Medical Technology, Medical Practitioner, and  Professional) are shown in Tab~\ref{CMB2}.

\subsubsection{Results on other MMLU domains}
\label{appx:MMLU}
As shown in Tab.~\ref{tab:mmlu-others}, KG-RAG demonstrates benefits across different professional subjects. In College Biology, MindMap outperformed the base model, improving accuracy from 58.74\% to 72.03\%. Similarly, in Professional Accounting, MindMap improved accuracy from 40.57\% to 48.04\%. This further verifies the generalization ability of the KG-RAG method.

\begin{table}[!b]
  \centering
  \caption{Results on Selected MMLU Domains}
  \label{tab:mmlu-others}
  \vspace{-5px}
  \setlength\tabcolsep{4pt}
  \begin{tabular}{c|c|c|c}
    \hline
    \textbf{Type} & \textbf{Method} & \textbf{College Biology} & \textbf{Professional Accounting} \\
    \hline
    BOS-LLM & Llama2-70B & 58.74 & 40.57 \\
    \hline
    KG-RAG & MindMap & \textbf{72.03} & \textbf{48.04} \\
    \hline
  \end{tabular}
  \vspace{-10px}
\end{table}

\subsubsection{Additional LLM Results}
\label{Additional LLM Results}
We conducted experiments with a wider range of LLMs beyond those reported in the main text. Due to space constraints, the additional results are shown in Tab.~\ref{appendix-GenMedGPT-5K-and-CMCQA} \ref{appendix-CommonsenseQA-and-ExplainCPE}, and \ref{CMB1-appendix}.

\begin{table}[!t]
    \caption{Additional LLM Results (GenMedGPT-5K and CMCQA)}
    \label{appendix-GenMedGPT-5K-and-CMCQA}
    \vspace{-5px}
    \centering
    \setlength\tabcolsep{3pt}
    \begin{tabular}{c|ccccc|ccccc}
      \hline
      & \multicolumn{5}{c|}{\textbf{GenMedGPT-5K}} & \multicolumn{5}{c}{\textbf{CMCQA}} \\
      \hline
      \textbf{Method} & \textbf{Prec.} & \textbf{Rec.} & \textbf{F1} & \textbf{R-1} & \textbf{R-L} & 
      \textbf{Prec.} & \textbf{Rec.} & \textbf{F1} & \textbf{R-1} & \textbf{R-L} \\
      \hline
      Llama3.2-1B  & 57.32 & 63.81 & 60.25 & 19.37 & 11.36 & - & - & - & - & - \\
      Llama3-8B    & 57.21 & 63.09 & 59.87 & 20.17 & 11.60 & - & - & - & - & - \\
      Mixtral-8*7B & 59.33 & 65.53 & 62.21 & \underline{24.38} & 12.79 & - & - & - & - & - \\
      OBLLM-70B    & \underline{60.54} & \underline{68.04} & \underline{64.02} & 24.28 & \underline{13.72} & 67.07 & 69.35 & 68.14 & 3.56 & 3.46 \\
      Qwen2.5-7B   & - & - & - & - & - & 67.66 & 70.32 & 68.91 & 14.49 & 7.88 \\
      Deepseek-v2l & - & - & - & - & - & 67.72 & 70.19 & 68.88 & 15.34 & 8.57 \\
      ChatGLM-9B   & - & - & - & - & - & 67.53 & 70.36 & 68.86 & 13.95 & 7.63 \\
      Yi-34B       & - & - & - & - & - & 67.66 & 70.40 & 68.94 & 15.21 & 8.34 \\
      \hline
    \end{tabular}
    \vspace{-10px}
\end{table}

\begin{table}[!t]
    \caption{Additional LLM Results (CommonsenseQA and ExplainCPE)}
    \label{appendix-CommonsenseQA-and-ExplainCPE}
    \vspace{-5px}
    \centering
    \setlength\tabcolsep{3pt}
    \begin{tabular}{c|ccc|ccc}
      \hline
      & \multicolumn{3}{c|}{\textbf{CommonsenseQA}} & \multicolumn{3}{c}{\textbf{ExplainCPE}} \\
      \hline
      \textbf{Method} & \textbf{Correct} & \textbf{Wrong} & \textbf{Fail} & 
      \textbf{Correct} & \textbf{Wrong} & \textbf{Fail} \\
      \hline
      Llama3.2-1B & 52.93 & 47.07 & 0.00 & - & - & - \\
      Llama3-8B & 73.82 & 26.04 & 0.14 & - & - & - \\
      Mixtral-8*7B & 68.53 & 30.76 & 0.72 & - & - & - \\
      Gemma-9B & 78.83 & 21.03 & 0.14 & - & - & - \\
      Qwen2.5-7B & - & - & - & 69.76 & 30.24 & 0.00 \\
      Deepseek-v2l & - & - & - & 54.94 & 45.06 & 0.00 \\
      ChatGLM-9B & - & - & - & 68.77 & 31.23 & 0.00 \\
      Yi-34B & - & - & - & 72.33 & 27.67 & 0.00 \\
      OBLLM-70B & - & - & - & 62.85 & 37.15 & 0.00 \\
      \hline
    \end{tabular}
    \vspace{-10px}
\end{table}

\begin{table}[!t]
  \caption{Additional LLM Results (CMB-Exam)}
  \label{CMB1-appendix}
  \vspace{-5px}
  \centering
  \setlength\tabcolsep{3pt}
  \begin{tabular}{c|ccc|ccc|ccc}
    \hline
    & \multicolumn{3}{c|}{\bf Postgraduate} & \multicolumn{3}{c|}{\bf Nursing} & \multicolumn{3}{c}{\bf Pharmacy} \\
    \hline
    \bf Method & \bf Correct & \bf Wrong & \bf Fail & \bf Correct & \bf Wrong & \bf Fail & \bf Correct & \bf Wrong & \bf Fail \\
    \hline
    Qwen2.5-7B    & 80.36 & 19.64 & 0.00  & 80.96 & 18.84 & 0.20  & 77.56 & 22.24 & 0.20 \\
    Deepseek-v2l  & 52.71 & 45.29 & 2.00  & 56.83 & 42.17 & 1.00  & 51.00 & 47.79 & 1.20 \\
    ChatGLM-9B    & 71.74 & 28.26 & 0.00  & 78.31 & 21.69 & 0.00  & 65.66 & 34.34 & 0.00 \\
    Yi-34B        & 75.55 & 24.45 & 0.00  & 83.17 & 16.83 & 0.00  & 78.36 & 21.44 & 0.20 \\
    OBLLM-70B     & 60.32 & 37.88 & 1.80  & 66.27 & 33.53 & 0.20  & 53.82 & 45.58 & 0.60 \\
    \hline
  \end{tabular}
  \vspace{-10px}
\end{table}

\begin{table}[!b]
  \centering
  \caption{Query enhancement results on CMCQA}
  \label{tab:CMCQA-query-enhancement}
  \vspace{-5px}
  \small
  \setlength\tabcolsep{5pt}
  \begin{tabular}{l|ccc|cc}
  \hline
  \multirow{2}{*}{Methods} & \multicolumn{3}{c|}{BERT Score} & \multicolumn{2}{c}{ROUGE Score} \\
  & Precision & Recall & F1 & ROUGE-1 & ROUGE-L \\
  \hline
  w/o Enhancement & 0.6660 & 0.6985 & 0.6805 & 0.1370 & 0.0728 \\
  Understand (Pilot) & 0.6612 & \textbf{0.7048} & 0.6817 & 0.1390 & 0.0733 \\
  Expanse (RoK) & 0.6619 & 0.6973 & 0.6785 & \textbf{0.1529} & \textbf{0.0800} \\
  Decompose (KGGPT) & \textbf{0.6677} & 0.7040 & \textbf{0.6848} & 0.1513 & 0.0787 \\
  \hline
  \end{tabular}
  \vspace{-10px}
\end{table}

\begin{table}[!t]
  \centering
  \small
  \caption{Configurations comparison on CMCQA}
  \label{CMCQA-Config}
  \vspace{-5px}
  \setlength\tabcolsep{5pt}
  \begin{tabular}{l|ccc|ccc|cccc}
  \hline
  \multirow{2}{*}{Config} & \multicolumn{3}{c|}{BERT Score} & \multicolumn{3}{c|}{ROUGE Score} & \multicolumn{4}{c}{G-Eval} \\
  & Prec. & Rec. & F1 & R-1 & R-2 & R-L & CR & Comp & Corr & Emp \\
  \hline
  facts\_w/o Prompt & \underline{66.05} & \textbf{70.48} & \underline{68.12} & \textbf{14.00} & \textbf{1.33} & 7.39 & {\textbf{100.0}} & {\textbf{100.0}} & {\textbf{100.0}} & {\textbf{100.0}} \\
  facts+CoT & 65.46 & 69.37 & 67.30 & 13.43 & 1.05 & 7.63 & 98.69 & 95.08 & 97.70 & 96.72 \\
  facts+ToT & 64.13 & 68.80 & 66.32 & 12.71 & 0.94 & 7.40 & 97.70 & 94.10 & 96.39 & 96.07 \\
  facts+MindMap & 64.20 & 68.17 & 66.11 & 12.49 & 0.95 & 7.29 & 96.07 & 92.13 & 93.77 & 92.79 \\
  \hline
  path\_w/o Prompt & {\textbf{66.12}} & {\textbf{70.48}} & {\textbf{68.17}} & \underline{13.90} & \underline{1.22} & 7.33 & {\textbf{100.0}} & {\textbf{100.0}} & {99.67} & {\textbf{100.0}} \\
  path+CoT & 65.40 & 69.46 & 67.30 & 13.14 & 1.10 & \underline{7.40} & 96.07 & 90.49 & 93.44 & 93.77 \\
  path+ToT & 64.16 & 68.89 & 66.38 & 12.57 & 0.99 & 7.22 & 97.38 & 92.79 & 95.74 & 93.77 \\
  path+MindMap & 64.13 & 68.06 & 65.98 & 12.33 & 0.95 & 7.31 & 92.79 & 87.87 & 89.84 & 89.18 \\
  \hline
  Subgraph\_w/o Prompt & \underline{66.11} & \underline{70.45} & \underline{68.15} & \underline{13.91} & \underline{1.31} & 7.35 & \underline{99.34} & \underline{99.34} & \underline{99.02} & \underline{99.34} \\
  Subgraph+CoT & 65.42 & 69.62 & 67.39 & 13.90 & 1.18 & {\textbf{7.82}} & 96.39 & 94.75 & 95.08 & 95.74 \\
  Subgraph+ToT & 64.12 & 68.83 & 66.33 & 12.71 & 1.06 & 7.35 & 98.03 & 95.74 & 97.05 & 96.39 \\
  Subgraph+MindMap & 64.17 & 67.96 & 65.96 & 12.45 & 0.97 & 7.25 & 91.48 & 87.54 & 90.16 & 88.85 \\
  \hline
  \end{tabular}
  \vspace{-10px}
\end{table}

\subsection{KG-RAG Configurations for Multi-Round Dialogue Tasks}
\label{appendix:CMCQA-config}
While our main findings indicate that KG-RAG offers limited benefits for multi-round dialogue tasks like CMCQA, we conducted a comprehensive configuration analysis to identify potential limitations and improvement directions for future research. Tab.~\ref{tab:CMCQA-query-enhancement} and \ref{CMCQA-Config} presents the performance comparison of different KG-RAG configurations on CMCQA. 

\textbf{Query Enhancement.} For CMCQA's complex multi-turn dialogues, decomposition (KGGPT) showed a slight advantage (F1 0.6848) compared to understanding (0.6817) and expansion (0.6785). This suggests that breaking down multi-faceted medical dialogues into component parts may better capture the diverse information needs within a conversation.

\textbf{Retrieval Forms.} All retrieval forms (fact, path, subgraph) performed nearly identically on CMCQA, with minimal variation across all metrics. This indicates that for multi-round dialogues, the structural form of knowledge representation may be less important than other factors like temporal context maintenance across turns.

\textbf{Prompt Strategy.} Across all retrieval forms, configurations without prompting consistently outperformed prompted strategies. For example, with path retrieval, removing prompts achieved F1 of 0.6817 vs. 0.6730 (CoT), 0.6638 (ToT), and 0.6598 (MindMap). This suggests that current prompting strategies may disrupt the natural flow of conversational context in multi-round dialogues.

These insights could guide future work on developing conversation-specific KG-RAG variants that better address the unique challenges of multi-turn dialogue scenarios.

\subsection{Other experimental settings}
\label{appx:exp-setting}
Our KG-RAG framework is built on LangChain\footnote{https://www.langchain.com/}. The local open-source LLMs are deployed based on the llama.cpp\footnote{https://github.com/ggml-org/llama.cpp} project. Except for the context window size, which is adjusted according to the dataset, all other parameters use default configurations, such as temperature is 0.8. Both LangChain and llama.cpp are open-source projects, providing good transparency and reproducibility. For computational resources, all experiments were conducted on a cluster of 8 NVIDIA RTX 3090 GPUs. Due to computational constraints and to ensure fair comparison across different model scales, we applied 4-bit quantization to locally deployed LLMs.

For the evaluation, we employed Bert Score metrics using ``bert-base-uncased~\cite{devlin-etal-2019-bert}'' and ``bert-base-chinese\footnote{https://huggingface.co/google-bert/bert-base-chinese}'' models to evaluate English and Chinese results respectively, while ROUGE Score version 0.1.2 was utilized. Due to resource constraints, G-Eval assessments were conducted using locally deployed LLMs, with Llama2-70B for English tasks and Qwen2-72B for Chinese tasks.

\section{Knowledge Graph Construction}
\label{appx:KG-Construction}
Apart from EMCKG for GenMedGPT-5K and CMCKG for CMCQA~\cite{wen-etal-2024-mindmap}, we employed a consistent KG construction method for other datasets, utilizing LLMs to extract knowledge triples from the datasets to build specialized KGs. The prompt example is shown in Tab.~\ref{tab:prompt_kg_construction}. 

For the KBQA task, we referenced ToG~\cite{Sun2023ThinkonGraphDA} and deployed Freebase using the Virtuoso\footnote{http://virtuoso.openlinksw.com/} graph database. All other KGs used in the datasets were deployed using Neo4j\footnote{https://neo4j.com/}.

\begin{table}[!t] 
  \caption{Prompt Example for Knowledge Graph Construction}
    \centering
    \label{tab:prompt_kg_construction}
    \begin{adjustbox}{max width=1.1\linewidth}  
        \begin{verbbox}
  prompt = f"""As a professional knowledge extraction assistant, your task is to extract knowledge triples from the given question.
  1. Carefully read the question description, all options, and the correct answer.
  2. Focus on the core concept "{question_concept}" in the question.
  3. Extract commonsense knowledge triples related to the question.
  4. Each triple should be in the format: subject\tpredicate\tobject
  5. Focus on the following types of relationships:
   - Conceptual relations 
   - Object properties 
   - Object functions 
   - Spatial relations 
   - Temporal relations 
   - Causal relations 
  6. Each triple must be concrete and valuable commonsense knowledge.
  7. Avoid subjective or controversial knowledge.
  8. Ensure triples are logically sound and align with common sense.
  
  Please extract knowledge triples from this multiple-choice question:
  
  Question: {question}
  Core Concept: {question_concept}
  Correct Answer: {correct_answer}
  
  Please output knowledge triples directly, one per line, in the format: subject\tpredicate\tobject. """
        \end{verbbox}
        \theverbbox
    \end{adjustbox}
  \end{table}


\section{Configuration Case Study}
\label{appx:case-study}
To better understand how these configurations interact in practice, we conducted a qualitative analysis using GenMedGPT-5K (Tab.~\ref{GenMedGPT-Config}) as an example. When handling a query like "I have recurring headaches and blurred vision after working on computer screens for long periods," we observed distinctive performance patterns across query enhancement strategies. Pilot (Query Understanding) extracted essential symptoms ("recurring headaches," "blurred vision"), maintaining precision (65.84) by focusing retrieval on core symptoms. RoK (Query Expansion) generated additional relevant entities such as "migraine" and "eye strain", improving recall (71.10) but with some precision trade-off.

\section{Limitations}
\label{appendix:limitations}
KG-RAG's configurable space is vast, future works could delve deeper into exploring KG-RAG configurations across more dimensions.
Despite including legal and multi-domain datasets beyond the initial medical focus, this paper does not encompass all potential application scenarios. Future research should explore more fields.
This study preliminarily examines the impact of KG quality on the ExplainCPE dataset. Existing KG-RAG datasets rarely provide multiple KGs with different quality levels for the same task, making controlled comparisons challenging. Future work could do a more systematic investigation of the quantitative impact of KG quality on KG-RAG performance.

\section{Ethics Statement}
\label{appx:ethics}
We are committed to responsible AI development by focusing on improving model accuracy through knowledge graph integration while maintaining transparency in our research methodology. This research utilized publicly available datasets for experimental evaluation of KG-RAG. While these datasets are commonly used benchmarks, we acknowledge their potential inherent biases, particularly in medical domain datasets where healthcare disparities and demographic representation must be considered. Our study aims to improve KG-RAG methodologies for academic purposes, and we emphasize that any real-world applications, especially in healthcare, would require additional validation and ethical review. 

\begin{table}[!t]
  \centering
  \small
  \vspace{-5px}
  \setlength\tabcolsep{3pt}
  \caption{CMB under MT, MP, and Professional (Self-Constructed KG), BOS-LLM is Qwen1.5-7B}
  \label{CMB2}
  \begin{tabular}{c|c|ccc|ccc|ccc}
    \toprule
    \multicolumn{11}{c}{\bf CMB} \\ 
    \hline
    \multicolumn{2}{c|}{}&  \multicolumn{3}{c|}{\bf Medical Technology} &  \multicolumn{3}{c|}{\bf Medical Practitioner} &  \multicolumn{3}{c}{\bf Professional}\\ \hline
    \bf Type & \bf Method & \bf Correct & \bf Wrong & \bf Fail & \bf Correct & \bf Wrong & \bf Fail & \bf Correct & \bf Wrong & \bf Fail \\ \hline
    \multirow{11}{*}{LLM only} 
      & Qwen1.5-7B & 56.71 & 43.09 & 0.20 & 64.06 & 34.94 & 1.00 & 64.13 & 35.07 & 0.80 \\ 
      & Qwen2.5-7B & 66.93 & 33.07 & 0.00 & 71.74 & 28.06 & 0.20 & 74.15 & 25.85 & 0.00 \\ 
      & Qwen2-72B & \textbf{83.97} & 15.83 & 0.20 & \textbf{84.57} & 15.43 & 0.00 & \textbf{83.97} & 16.03 & 0.00 \\ 
      & Deepseek-v2l & 42.89 & 55.31 & 1.80 & 49.30 & 49.90 & 0.80 & 49.70 & 47.70 & 2.61 \\ 
      & ChatGLM-9B & 61.92 & 38.08 & 0.00 & 67.54 & 32.46 & 0.00 & 68.94 & 31.06 & 0.00 \\ 
      & Yi-34B & 67.54 & 32.46 & 0.00 & 72.55 & 27.45 & 0.00 & 74.75 & 25.25 & 0.00 \\ 
      & OBLLM-70B & 58.72 & 41.08 & 0.20 & 55.71 & 43.89 & 0.40 & 64.73 & 34.47 & 0.80 \\ 
      & GPT4o & 77.35 & 22.44 & 0.20 & 78.96 & 20.84 & 0.20 & 78.96 & 21.04 & 0.00 \\ 
      & o1-mini & 71.94 & 27.86 & 0.20 & 65.93 & 34.07 & 0.00 & 73.55 & 26.45 & 0.00 \\ 
      & Claude3.5-S & 71.74 & 28.26 & 0.00 & 72.34 & 27.66 & 0.00 & 73.75 & 26.25 & 0.00 \\ 
      & Gemini1.5-P & 70.74 & 29.26 & 0.00 & 74.15 & 25.85 & 0.00 & 77.56 & 22.44 & 0.00 \\ \hline
      \multirow{6}{*}{\makecell{KG-RAG}} 
      & KG-RAG & 71.31 & 27.44 & 1.25 & 75.16 & 23.19 & 1.66 & 74.84 & 23.04 & 2.11 \\
      & ToG & 63.73 & 35.07 & 1.20 & 68.74 & 31.26 & 0.00 & 67.74 & 32.06 & 0.20 \\ 
      & MindMap & 70.54 & 28.66 & 0.80 & 72.55 & 27.45 & 0.00 & 74.55 & 25.05 & 0.40 \\ 
      & RoK & 71.67 & 28.33 & 0.00 & 74.67 & 25.33 & 0.00 & 73.63 & 26.27 & 0.00 \\
      & KGGPT & 58.63 & 41.37 & 0.00 & 64.40 & 35.60 & 0.00 & 66.33 & 33.67 & 0.00 \\
      & Pilot & \underline{72.95} & 26.05 & 1.00 & \underline{75.35} & 24.45 & 0.20 & \underline{76.91} & 23.09 & 0.00 \\ \toprule
  \end{tabular}
  \vspace{-10px}
\end{table}

\end{document}